\documentclass[runningheads]{llncs}
\usepackage{graphicx}
\usepackage{amsmath}
\usepackage{xcolor}

\DeclareMathOperator{\Tr}{Tr}
\DeclareMathOperator{\sbjto}{s.t.}
\DeclareMathOperator{\diag}{diag}

\newcommand{\x}{\mathbf{x}}

\newcommand{\h}{\mathbf{h}}
\newcommand{\Hid}{\mathbf{H}}
\newcommand{\I}{\mathbf{I}}
\newcommand{\E}{\mathbf{E}}
\newcommand{\X}{\mathbf{X}}
\newcommand{\W}{\mathbf{W}}

\begin{document}
\title{A Neural Network for Semi-Supervised Learning on Manifolds}
%
%
\author{Alexander Genkin*\inst{1} 
\and Anirvan M. Sengupta\inst{2,3} 
\and Dmitri Chklovskii\inst{1,2} 
} 

\institute{%
	Neuroscience Institute, NYU Langone Medical Center, 435 E 30th St, New York, NY 10016 USA \email{alexander.genkin@nyumc.org}\\
	\and Flatiron Institute, Simons Foundation, 162 5th Ave. New York, NY 10010 USA
    \email{dchklovskii@flatironinstitute.org}
	\and Dept.~of Physics and Astronomy, Rutgers University 136 Frelinghuysen Rd, Piscataway, NJ 08854 USA \email{anirvans@physics.rutgers.edu}
} 
\authorrunning{A. Genkin et al.}
%
%
\maketitle              
\begin{abstract}
Semi-supervised learning algorithms typically construct a weighted graph of data points to represent a manifold. However, an explicit graph representation is problematic for neural networks operating in the online setting. Here, we propose a feed-forward neural network capable of semi-supervised learning on manifolds without using an explicit graph representation. Our algorithm uses channels that represent localities on the manifold such that correlations between channels represent manifold structure. The proposed neural network has two layers. The first layer learns to build a representation of low-dimensional manifolds in the input data as proposed recently in \cite{Bumps}. The second learns to classify data using both occasional supervision and similarity of the manifold representation of the data. The channel carrying label information for the second layer is assumed to be ``silent" most of the time. Learning in both layers is Hebbian, making our network design biologically plausible. We experimentally demonstrate the effect of semi-supervised learning on non-trivial manifolds.

\end{abstract}

\section{Introduction}
 When labeled data are scarce or expensive to obtain, we often resort to semi-supervised learning which exploits the abundance of unlabeled data. For data concentrating on a lower dimensional manifold, it is often reasonable to assume smoothness, i.e., that data points adjacent on the manifold are likely to have similar values of the target variable (the label). Then, learning the manifold structure from both labeled and unlabeled data can assist in label prediction. \cite{ando_zhang,belkin2006manifold,bengio200611,zhu2003semi}.

In machine learning, an online method updates the model incrementally as it receives training data in a sequential manner. This approach is to be contrasted with offline machine learning, which generates the best model by learning on the entire training data set at once. Online learning is used either because it is computationally infeasible to train over the entire dataset, or it is used where the algorithm has to dynamically adapt to new patterns in the data, e.g. when the data itself is generated in real time. The last situation is particularly relevant in the context of neuronal networks.

Our brains likely rely on online semi-supervised learning to generate behavior. As our sensory organs stream data about the world they are analyzed in real time to produce behaviorally relevant output. While most of the sensory data lack labels, some supervision is available from other sources such as inter-personal communication.   

To represent a data manifold, semi-supervised learning algorithms typically construct an adjacency graph whose vertices are labeled and unlabeled data points and edge weights represent their adjacency on the manifold. However, such representation is impractical in the online setting where the data are streamed sequentially and the labels are predicted on the fly. Furthermore, the online setting does not have the memory capacity to store the past data. 

Thus, there is a need for online semi-supervised algorithms both for modeling neural computation and solving general machine learning tasks. Whereas existing online algorithms \cite{goldberg2008online} can rely on a sparse representation, they still require memory quadratic in the dimensionality of data. In addition, these algorithms rely on the availability of an adjacency measure between new and stored data points.

In this paper, we propose a biologically plausible neural network for online semi-supervised learning (Figure \ref{fig:bumps}, left). By avoiding explicit representation of the adjacency graph our network can process unlimited-size datasets in online setting. Moreover, as required by biology, the network relies only on local learning rules meaning that synaptic weight update depends on the activity of only the two neurons this synapse connects. 

The network has two layers. The first layer learns the manifold structure of the data by representing each datum as a sparse vector whose components represent overlapping localities on the manifold. The manifold structure is captured by the correlations between the components carried by corresponding channels. Because most existing algorithms for sparse representations, such as  \cite{LCC}, do not have natural neural implementations we base our work on the recently developed manifold tiling algorithm \cite{Bumps}. Inspiration for such design comes from biological neural networks such as place cells in the rodent hippocampus. 

The second layer learns a classifier using both occasional supervision and the similarity of the manifold representation of the data provided by the first layer. In our neural network, the supervision signal is not fed back from downstream layers of the network like in perceptron or back-propagation networks, but comes along and synchronously with the data from the previous layer. To make it semi- (rather than fully) supervised, the label signal is assumed to be ``silent" most of the time.
The output attempts to predict the correct label when that signal is not available, otherwise it just reproduces the label.

We derive both the activity dynamics and the learning rules in each layer from the principle of similarity preservation \cite{pehlevan2015MDS} which was previously used in the unsupervised setting. Starting with a similarity preserving objective function allows us to analyze the output of the algorithm and obtain biologically plausible local learning rules. 


We demonstrate experimentally the effectiveness of this semi-supervised network compared to fully supervised online learning. Moreover, we observe that online semi-supervised learning may be competitive with offline methods, especially on smaller samples. This is an important advantage allowing our network to adapt quickly when the manifold shape or the labels are changing with time.


\section{Review of the Manifold-Tiling Network Derived from Non-negative Similarity Matching }

\begin{figure}[t]\centering
    \includegraphics[width=.4\textwidth,height=3.7cm]{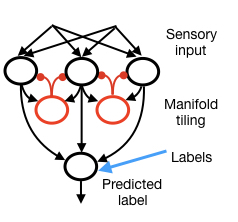} 
    \raisebox{-0.1\height}{
    \includegraphics[width=.55\textwidth,height=4.95cm]{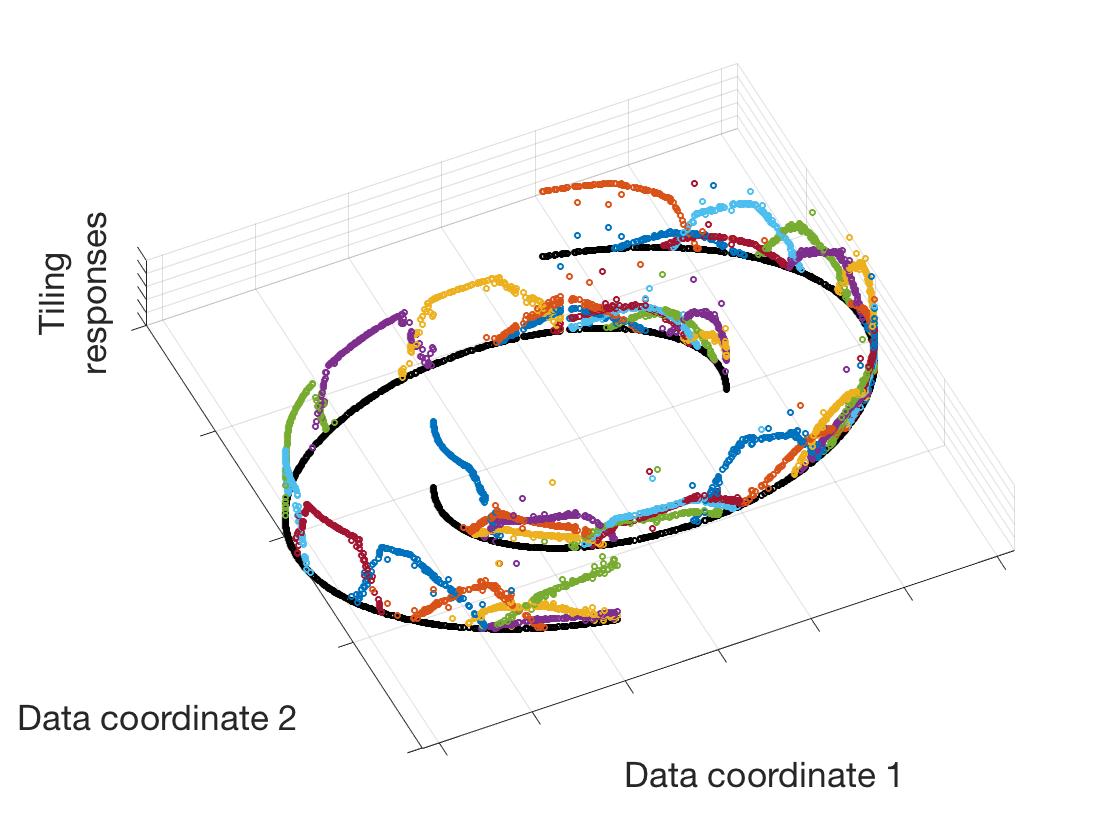}} 
    \caption{\label{fig:bumps} 
    {\it Left:} Two-layer network  for semi-supervised learning. The first layer learns  manifolds (the upper layer with interneurons in red),  the second layer 
    is one neuron that learns a classifier in the semi-supervised  manner. The output of the network predicts the label value when the labels channel is silent, otherwise it reproduces the label.
    {\it Right:} Receptive fields of manifold tiling. Data are 2000 points sampled from two arcs in 2D (black dots); in the third dimension we show responses of individual channels to the corresponding data points, each channel is assigned a color. 
    }
\end{figure}

To introduce our notation, let the input to the network be a set of vectors, $\x_t\in  R^n, t=1,\ldots,T$, coming from $n$ channels at time $t$. In response, the manifold learning network layer outputs an activity vector, $\h_t\in R^m, t=1,\ldots,T$, $m$ being the number of output channels, or hidden units in our two-layer network, Figure \ref{fig:bumps}, left. 

Manifold-tiling networks have been derived \cite{Bumps} from similarity-preserving objectives \cite{NIPS2015_5885} with a non-negativity constraint. Similarity preservation postulates that similar input pairs, $\x_t$ and $\x_{t'}$, evoke similar output pairs, $\h_t$ and $\h_{t'}$. Similarity of a pair of vectors can be quantified by their scalar product. Nonlinear manifolds can be learned by constraining the sign of the output and introducing a similarity threshold $\alpha$. \cite{Bumps} propose an optimization problem: 
\begin{align}
\label{eq:nsm1a}
&\min_{\substack{ \Hid\ge 0 \\ \diag{\Hid^\top\Hid} \le \I }} -\Tr((\X^\top\X-\alpha \E)\Hid^\top\Hid) \\ \nonumber
&=\min_{\h_t\ge 0,\; \|\h_t\|_2^2\le1
} -\sum_{t,t'}(\x_t^\top \x_{t'}-\alpha)\h_t^\top \h_{t'}.
\end{align}
Here matrix notation was introduced: $\X\equiv[\x_1,\ldots,\x_T]\in    R^{n\times T}$ and $\Hid\equiv[\h_1,\ldots,\h_T]\in R^{m\times T}$, and ${\bf E}$ is a matrix of all ones.

Intuitively, \eqref{eq:nsm1a} attempts to preserve similarity for similar pairs of input samples but orthogonalizes the outputs corresponding to dissimilar input pairs. Indeed, if the input similarity of a pair of samples $t,t'$ is above a specified threshold, $\x_t^\top \x_{t'}>\alpha$, then the output vectors $\h_t$ and $\h_{t'}$ would prefer to have $\h_t^\top \h_{t'} \approx \x_t^\top \x_{t'}-\alpha$, i.e., they would be similar. If, however, $\x_t^\top \x_{t'}<\alpha$, then they would tend to be orthogonal, $\h_t^\top \h_{t'}=0$, since the lowest value of $\h_t^\top \h_{t'}$ for $\h_t,\h_{t'}\ge 0$ is zero. As $\h_t$ and $\h_{t'}$ are nonnegative, to achieve orthogonality, the output activity patterns for dissimilar patterns would have non-overlapping sets of active output channels.   In the context of manifold representation, \eqref{eq:nsm1a} strives to preserve in the $\h$-representation the local geometry of the input data cloud in $\x$-space and let the global geometry emerge out of the nonlinear optimization process.

Fig.~\ref{fig:bumps} illustrates manifold tiling on a two spiral arcs in two dimensions, showing the receptive fields of output channels in the third dimension. Receptive fields tile the arcs with overlaps, but there is no overlap between separate arcs.

To derive a neural network that optimizes \eqref{eq:nsm1a}, we express the norm constraint in the Lagrangian form:
\begin{align}\label{lagrange}
\min_{\forall t:\, \h_t\ge 0 } \, \max_{\forall t:\, {\bf u}_t\ge 0} & -\frac 1T \sum_{t,t'}(\x_t^\top \x_{t'}-\alpha)\h_t^\top \h_{t'} \\ \nonumber
&+\sum_t{\bf u}_t^\top{\bf u}_{t}(1-\h_t^\top\h_{t}).
\end{align}
Here, unconventionally, the non-negative Lagrange multipliers that impose the inequality constraints are factorized into inner products of two non-negative vectors (${\bf u}_t^\top{\bf u}_{t}$). In the second step, we introduce auxiliary variables, $\W, {\bf b}, {\bf V}_t$ \cite{pehlevan2018similarity}: 
%
%
\begin{align}\label{Online1}
&\min_{\forall t: \, \h_t\ge 0 } \, \max_{\forall t: \, {\bf u}_t\ge 0} \min_{\W} \max_{\bf b}\min_{\forall t:\, {\bf V}_t\geq 0} \; T\Tr (\W^\top \W)
- T\| {\bf b} \|_2^2 +  \\ \nonumber
&\qquad  + \sum_t \Bigl( -2\x_t \W^\top \h_t +2\sqrt{\alpha}  \h_t^\top {\bf b}+  \|{\bf u}_t \|_2^2 -  
2 {\bf u}_t{\bf V}_t\h_t + \Tr ({\bf V}_t^\top {\bf V}_t) \Bigr)
\end{align}
The equivalence of \eqref{Online1} to \eqref{lagrange} can be seen by performing the $\W, {\bf b}$, and ${\bf V}_t$ optimizations explicitly and plugging back in the optimal values. Eq.~\eqref{Online1} suggests a two-step online algorithm (see \cite{Bumps} for full derivation). For each input $\x_t$, in the first step, one solves for $\h_t$, ${\bf u}_t$ and ${\bf V}_t$, by projected gradient descent-ascent-descent,
\begin{align} 
\left [ \begin{array}{c} {\bf h}_t \\ {\bf u}_t \\ {\bf V}_t \end{array}\right] \longleftarrow \left [ \begin{array}{c} {\bf h}_t + \gamma_h \left(\W\x_t-{\bf V}_t^\top {\bf u}_t -\sqrt{\alpha}{\bf b}\right)\\ {\bf u}_t + \gamma_u\left(-{\bf u}_t+{\bf V}_t{\bf h}_t\right) \\ {\bf V}_t + \gamma_V \left({\bf u}_t\h_t^\top -{\bf V}_t\right)\end{array}\right]_+,
\end{align}
where $\gamma_{h,u,V}$ are step sizes. This iteration can be interpreted as the dynamics of a biologically plausible neural circuit (Fig.~\ref{fig:bumps}, right, the upper layer), where components of $\h_t$ are activities of excitatory neurons, ${\bf b}$ is a bias term, components of ${\bf u}_t$ are activities of inhibitory neurons (shown in red), and $\W$ is the feedforward connectivity matrix. ${\bf V}_t$ is the synaptic weight matrix from excitatory to inhibitory neurons, which undergoes a fast time-scale anti-Hebbian plasticity, which in computer simulation means repeated updates within one $t$ step. In the second step, $\W$ and ${\bf b}$ are updated by gradient descent-ascent:
\begin{align*}
\W \leftarrow \W + \eta\left(\h_t\x_t^\top - \W \right), &&
{\bf b} \leftarrow {\bf b} + \eta\left(\sqrt{\alpha}\h_t -{\bf b}\right),
\end{align*}
where $\W$ is going through a slow time-scale Hebbian plasticity and ${\bf b}$ through homeostatic plasticity. The parameter $\eta$ is a learning rate. 



\section{A Neural Network for Semi-Supervised Learning}

In this section, we propose a neural network architecture for semi-supervised learning. In our approach, contrary to the widely accepted schemes, the label signal is not fed back from downstream layers of the network but comes along and synchronously with the rest of the data. To make it semi- (rather than fully) supervised, the signal is assumed to be ``silent" most of the time. 


Consider a  classification problem with the input stream of data, $\lbrace{\bf h}_1,\ldots,{\bf h}_t,\ldots\rbrace$, where ${\bf h}_t \in R^m$, and the corresponding class labels $\lbrace\tilde{z}_1,\ldots,\tilde{z}_t,\ldots\rbrace$, where in a binary case $\tilde{z}_t \in \{-1,+1\}$. The labels are occasionally signalled by a channel carrying values $z_t=\theta_t \tilde{z}_t$, where $\theta_t \in \{0,1\}$ either masks or reveals the true label. 
The data from the previous layer and the label channel are combined in the semi-supervised learning neuron, Fig. \ref{fig:bumps}, right, bottom layer.

Consider a time period of $1,\ldots,T$, where the inputs are organized into a matrix ${\bf H}=[{\bf h}_1,\ldots,{\bf h}_T]$ and a vector of (partly hidden) labels ${\bf z}^\top=(z_1,\ldots,z_T)$. The output ${\bf y}^\top=(y_1,\ldots,y_T)$ needs to reproduce the label signal, so we employ a quadratic loss function $\| {\bf y}-{\bf z}\|^2$. We express the assumption of smoothness of predicted label ${\bf y}$ on the manifold using the similarity alignment \cite{pehlevan2018similarity} between the input and output Gramians: $\Tr ({\bf H}^\top {\bf H}  {\bf y} {\bf y}^\top$). Also, as the label only takes values $1$ and $-1$, we restrict the output to stay within those limits. This gives rise to the following optimization problem:
\begin{align} \label{eq:zneuronOffline2}
&\min_{\bf y}\|{\bf y} - {\bf z}\|^2 - \frac{\mu}{T} \Tr ({\bf H}^\top {\bf H}  {\bf y} {\bf y}^\top ), \\ \nonumber 
  &\sbjto \;\;\;  -1 \le y_t \le 1, \;\; t=1,\ldots,T,
\end{align}
where we also introduced a regularization coefficient $\mu$ controlling the relative importance of the two parts of the objective function.

To derive an online algorithm, following \cite{pehlevan2018similarity}, we introduce an auxiliary variable ${\bf w}$ and expand in time:
\begin{align} \label{eq:zneuronONline2}
&\min_{\bf y} \min_{\bf w}  \frac{1}{T} \sum_t \Big[ \frac{1}{2} (y_t - z_t)^2  
  - \mu y_t {\bf w}^\top {\bf h}_t  \Big] + \frac{\mu}{2} {\bf w}^\top {\bf w} 
\\ \nonumber 
  &\sbjto \;\;\;  -1 \le y_t \le 1, \;\; t=1\ldots T.
\end{align}
Optimizing over ${\bf w}$, we obtain: ${\bf w} = \frac{1}{T}\sum_t y_t {\bf h}_t$, which makes it clear the new formulation is equivalent to Eq. \eqref{eq:zneuronOffline2}. The advantage of this formulation is that it suggests a two-step online algorithm. For each input ${\bf h}_t$, on the first step, one solves for the instantaneous output $y_t$ under fixed ${\bf w}$:
\begin{equation} \label{eq:ystep}
   y_t = \max(-1,\min(1, \mu {\bf w}_t^\top {\bf h}_t + z_t ))
\end{equation}
On the second step, ${\bf w}$ is updated as:
\begin{equation} \label{eq:wstep}
   {\bf w} \leftarrow \frac{t}{t+1} {\bf w} + \frac{1}{t+1} y_{t} {\bf h}_{t} 
\end{equation}

This also maps well onto a biologically plausible neural network where components of ${\bf w}$ are interpreted as synapse weights, updated by local Hebbian rule. We assume that the  synapse weight of the $z$ channel is not changing, thus differentiating it from the other input channels. We set this weight to be equal to 1 without loss of generality. The algorithm is initialized with ${\bf w}=0$, assuming no prior information. 

An alternative objective function can be obtained by expressing the loss as $-{\bf y}^\top{\bf z}$ and adding an entropy-like regularizer treating $(y_t+1)/2$ as a probability estimate for $\tilde{z}_t=1$:
\begin{align} \label{eq:zneuronOfflineEntropy}
\min_{\bf y}  &-{\bf y}^\top{\bf z} -  \frac{\mu}{2T}  \Tr ({\bf H}^\top {\bf H}  {\bf y} {\bf y}^\top ) \\ \nonumber
&- \sum_t \Big[ \frac{1+y_t}{2} \log(\frac{1+y_t}{2}) + \frac{1-y_t}{2} \log(\frac{1-y_t}{2}) \Big] 
\end{align}
The solution of this optimization problem is the familiar sigmoidal neuron rule: 
\begin{equation} \label{eq:ytanh}
   y_t = \tanh{( \mu {\bf w}_t^\top {\bf h}_t + z_t ))}
\end{equation}
with the same update for ${\bf w}$ as in Eq.~\eqref{eq:wstep}. The behavior of both algorithms is almost indistinguishable, so we only report the results from Eq.~\eqref{eq:ystep}. 
%
%
%
\section{Numerical Experiments}

\begin{figure}[ht]\centering
\includegraphics[width=.5\textwidth]{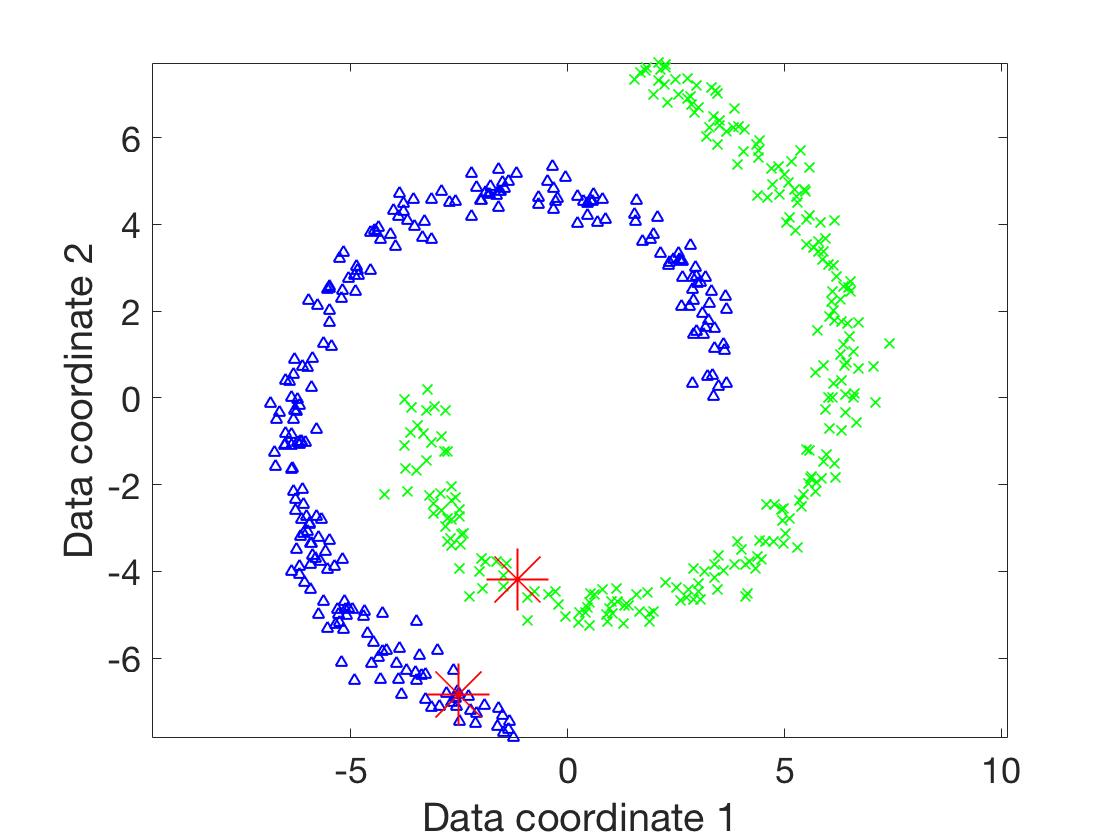}
\\
\includegraphics[width=.55\textwidth,height=4.5cm]{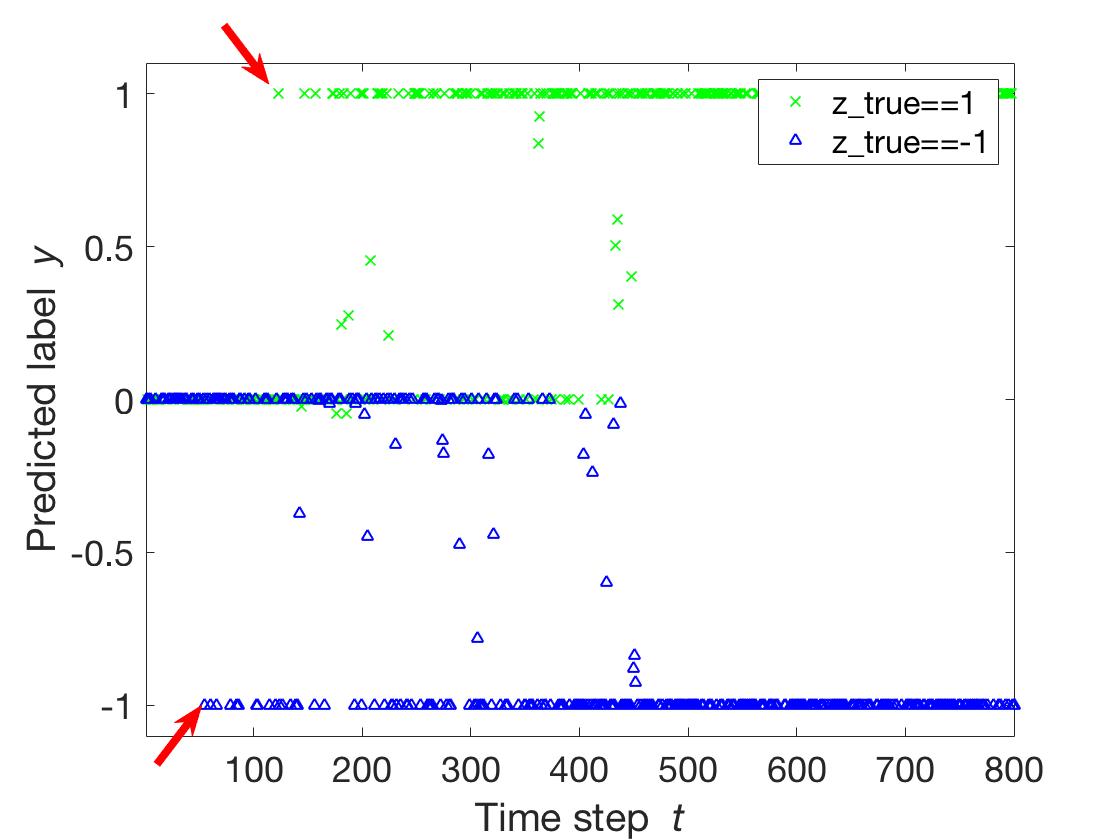}  \raisebox{0.1\height}{\includegraphics[width=.44\textwidth]{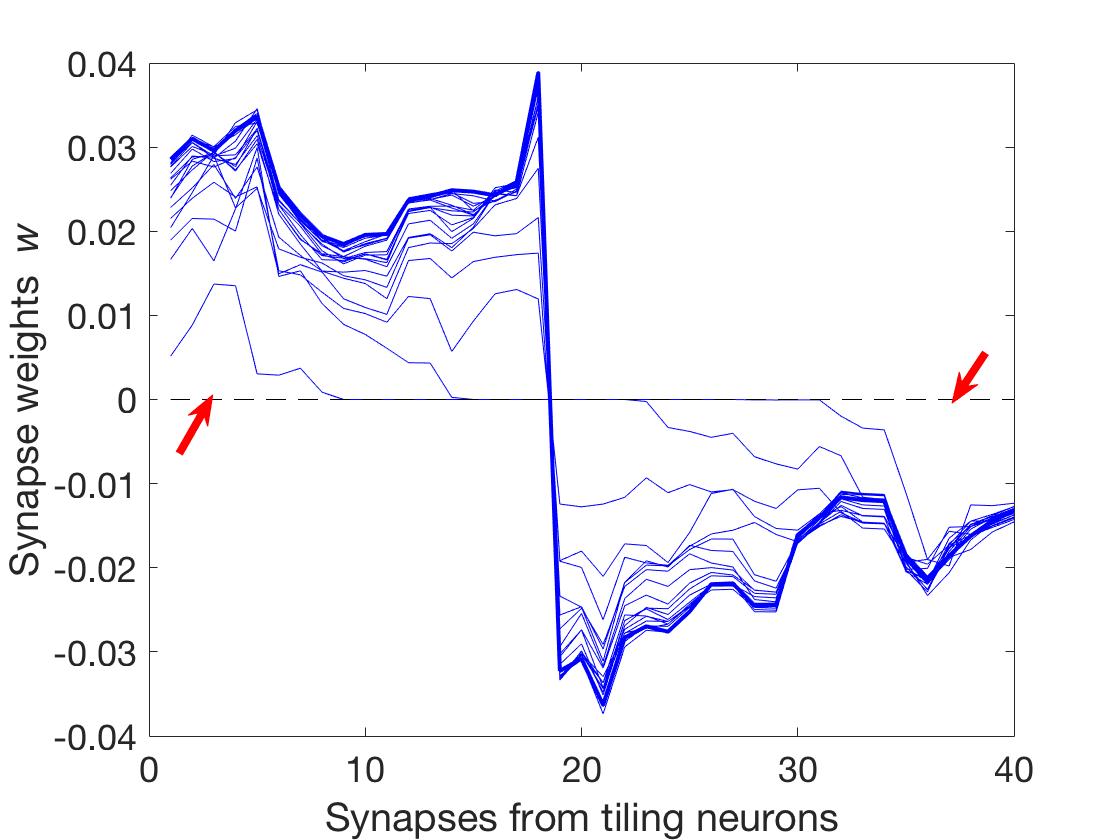}}
\caption{\label{fig:2spirals_zneuron} Semi-supervised learning on the ``two moons" dataset with two labeled points. 
\textit{Top:} ``Two moons" in 2D, classes - green crosses and blue triangles; red asterisks indicate two labeled points. 
\textit{Bottom left:} Label predicted by the semi-supervised neuron ($y_t$ values) in time, green crosses and blue triangles as the true class of the input. Labeled points indicated by arrows.   
\textit{Bottom right:} Propagation of labels is shown by the weights of the tiles in time. Tiles are ordered along the $x$ axis according to their locations on the ``moons": ``crosses" on the left and ``triangles" - on the right. Lines show temporal dynamics of the weights, and only every 100\textsuperscript{th} time point is shown. Arrows indicate tiles where labeled points fall. 
}
\end{figure}

We apply our algorithm to the 
synthetic dataset designed as ``two moons": two classes are sets of points in 2D, each concentrated around a spiral arc, Fig.~\ref{fig:2spirals_zneuron}, top. Such a synthetic dataset is widely used as a test for semi-supervised learning algorithms (see, e.g., \cite{belkin2006manifold,goldberg2008online}). Note that the classes are not linearly separable, and can be separated only when their manifold structure is discovered. Upon discovering the manifold structure, intuitively, the data can be classified using only one labeled example for each class, see red asterisks in Fig.~\ref{fig:2spirals_zneuron}, top.

Our network solves this highly non-linear classification problem. The first layer learns units that tile each ``moon" with overlaps while no unit is shared between the two moons. The second layer propagates labeling information along links formed by correlations in the tile responses. We generated data points randomly and uniformly, only placing two labeled points early in the data stream. 
We used tiling layer with 40 neurons and semi-supervised neuron with $\mu=1000$. The Fig.~\ref{fig:2spirals_zneuron}, 
bottom row,
illustrates the working of the semi-supervised neuron. As seen on  Fig.~\ref{fig:2spirals_zneuron}, bottom left, the output is zero until labeled points arrive. Then there is a transition period, during which the label signals propagate along correlated tiles. Finally, the responses stabilize to correct values: 1 for green, -1 for blue. Fig.~\ref{fig:2spirals_zneuron}, bottom right, illustrates propagation of the labeling information. Initially, all weights are zero. When a labeled point arrives, weights corresponding to the tiles overlapping this point increase in absolute value. That signal gradually propagates until all synapses corresponding to ``green moon" get positive weights and, all ``blue" ones - negative weights.

\begin{figure}[ht]\centering
    \includegraphics[width=.6\textwidth]{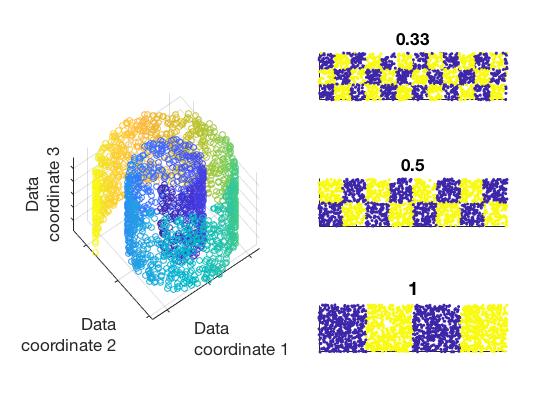} 
    \\
    \includegraphics[width=.49\textwidth]{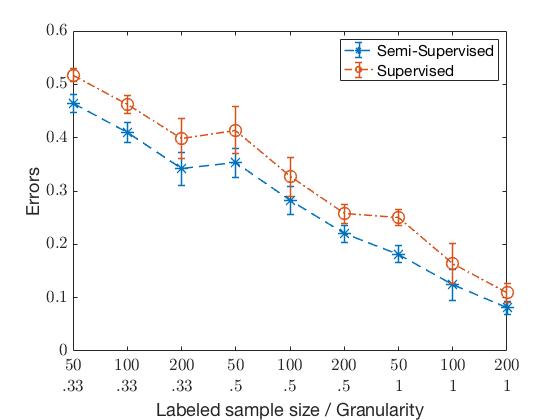}
    \includegraphics[width=.49\textwidth]{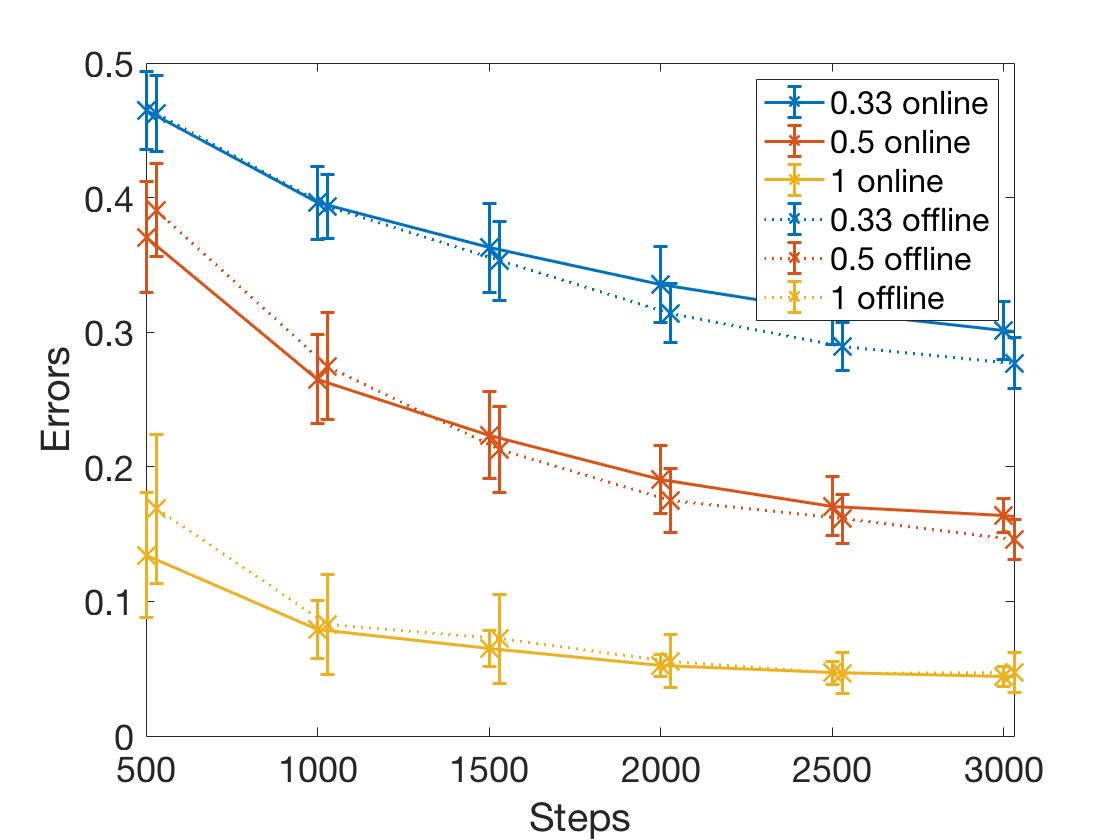}
  
    \caption{\label{fig:Chessboard} \textit{Left:} 3D Swiss roll manifold and three binary classification problems on the unrolled manifold. Granularity of the chessboard decreases from top to bottom. \textit{Center:}  Semi- vs. fully supervised: comparison of our semi-supervised algorithm with a fully supervised algorithm in the online setting.
    \textit{Right:} Online vs. offline: comparison of our online semi-supervised learning algorithm with the offline semi-supervised  algorithm (see text) for different granularity levels of the chessboard. }
\end{figure}

Next, we apply our network to a larger dataset, a 3D Chessboard on a Swiss roll, Fig.~\ref{fig:Chessboard}, left. All the data live on a 2D Swiss roll manifold and the two classes are defined by the squares of the chessboards. We consider chessboards with varying square sizes with the most fine-grained chessboard being most difficult for classification. Whereas linear classifiers per se can not solve this problem, after learning the manifold classification is linear.

We compare our semi-supervised algorithm with an online fully supervised classifier -- logistic regression. 
Both algorithms get the same input stream of 2000 data points, of which 50, 100, or 200 randomly selected are labeled and the rest are unlabeled.
The input for both classification algorithms is the output of tiling with 200 neurons. Parameters $\mu$ for our neuron and learning rate for logistic regression are selected for best results of each algorithm.  All runs repeated 10 times to obtain error bars.

Both algorithms classify each input using their current weights. However, the fully supervised algorithm cannot update its weights when an unlabeled example arrives, unlike the semi-supervised algorithm. Indeed, experiments show that the semi-supervised neural network performs better than the supervised classifier (Fig.~\ref{fig:Chessboard}, center), demonstrating its ability to take advantage of unlabeled data.

Next, we compare our online algorithm with an offline  semi-supervised learning algorithm. For the latter, we use a state of the art linear SVM with Laplacian penalty following \cite{belkin2006manifold}, but with a twist \footnote{In a separate experiment we made sure this twist only improves the results}: for the linear case we assume smoothness of weights rather than labels. This means that components of the separating vector {\bf w} should have similar values when corresponding tiling components have highly overlapping receptive fields. The degree of overlap between  receptive fields can be measured by dot products between tiling components, and can be calculated on both labeled and unlabeled data. Then the Gramian $S=\frac{1}{T}{\bf H}{\bf H}^\top$ can be thought of as the adjacency matrix of a graph where vertices are tiling components. The graph will be fairly sparse due to the nature of tiling. The graph Laplacian penalty is then:
\begin{equation} \label{grLap}
    \sum_{i,j} (w_i-w_j)^2 S_{i,j} = 2 {\bf w}^\top L {\bf w}, \;\;\;
        {\rm where} \;\;  L= {\rm  diag}(S{\bf 1})-S,
\end{equation}
and the objective function of linear SVM with Laplacian penalty takes the form:
\begin{equation}  \label{svmLap}
  \min_{{\bf w},b}  \sum_t [1-z_t({\bf w}^\top {\bf h}_t + b)]_+ + \lambda \|{\bf w}\|^2
     + \mu {\bf w}^\top L {\bf w}
\end{equation}
where the index $t$ runs through the labeled samples only, with $z_t \in \{-1,1\}$ being labels. 

In this experiment, the online algorithm is fed a data stream with 0.05\% of samples randomly labeled. Then at every 500th step the classification rule obtained up to this point is applied to a separate test set of 2000 samples. At the same step, the SVM with Laplacian regularization, Eq.~\eqref{svmLap}, is trained on all data seen online so far and tested against the same test set. 
As before, the input for both algorithms is the output of tiling with 200 neurons. Parameters $\mu$ for our neuron and learning rate for logistic regression are selected for best results of each algorithm.  All runs repeated 10 times to obtain error bars.

Offline algorithm has an advantage of considering all data samples before taking decision on labeling, while online algorithm has to assign a label estimate to each data sample as it appears. Results on Fig. \ref{fig:Chessboard}, right, show, however, that with enough smoothness (i.e., coarser granularity in the ``Chessboard" example), the online algorithm perform closely to the offline one. Moreover, online algorithm can perform better than the offline one while the number of presented data points is small (e.g., less than approximately 1200 with granularity 0.5). But small sample sizes is exactly the situation where semi-supervised learning is supposed to be helpful. The ability of the online algorithm to adapt quickly is also important when there is a drift in the manifold shape or the labels.

%
\section{Relation to Graph Laplacian}

\begin{figure}\centering
\includegraphics[width=0.49\textwidth]{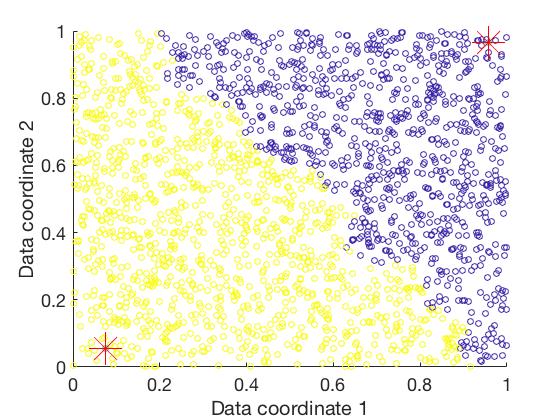} \includegraphics[width=0.49\textwidth]{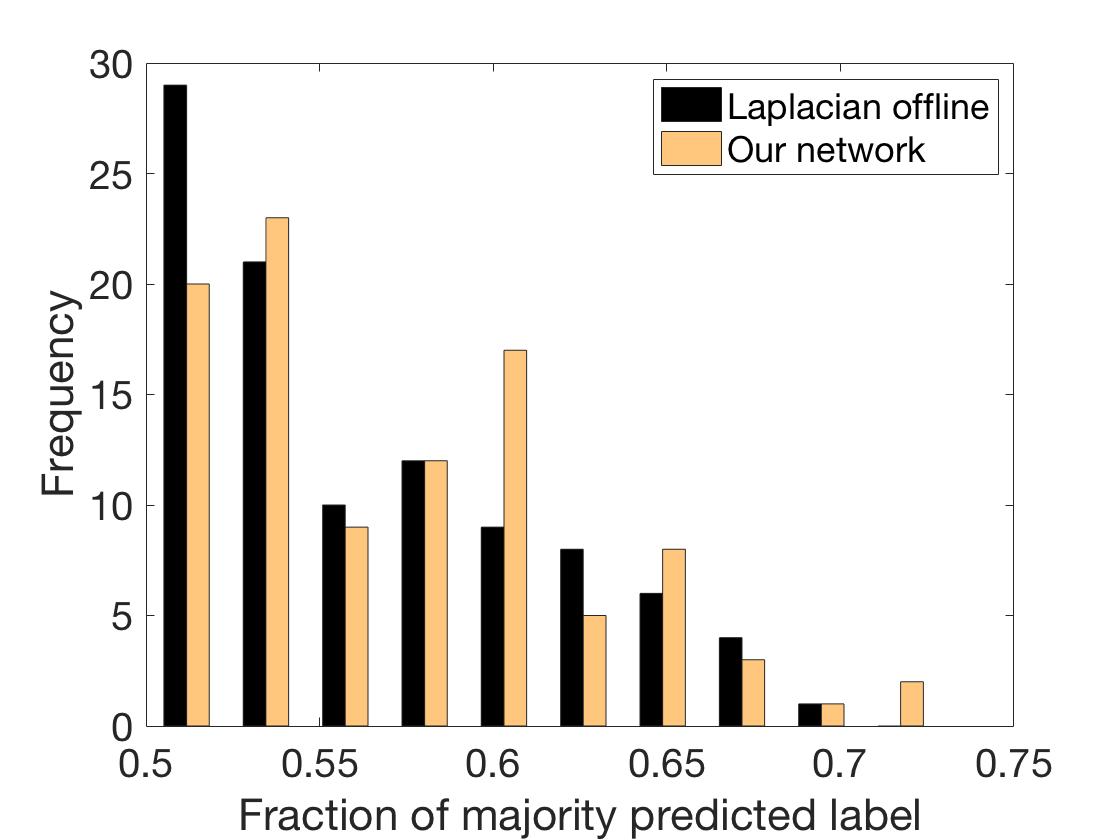} 
\caption{\label{fig:noMinCut} {\it Left:} Linear classification with Laplacian regularization, points colored by predicted label; labeled points are red arrows. {\it Right:} The histogram of 100 repetitions of simulation of our measure of imbalance, namely, the fraction of the majority predicted label for Laplacian solution offline and for our network. Note that the frequencies fall off in both methods as the fraction moves up from 0.5. 
}
\end{figure}

Existing algorithms for semi-supervised learning on manifolds typically utilize the graph Laplacian for smoothness regularization \cite{belkin2006manifold,zhu2003semi,bengio200611}, see the last term in Eq.~\eqref{svmLap}. This follows from the analysis of \cite{shi2000normalized}, which showed that graph Laplacian regularization results in classifier corresponding to normalized graph cut, which helps avoid heavily imbalanced classes. In contrast, our smoothness term, last term, in Eq.~\eqref{eq:zneuronONline2}, lacks diagonal normalization of Laplacian. When optimized exactly, it should lead to the minimum cut of the graph, which is prone to generate classes of very different size \cite{shi2000normalized}. Consider a simple example of a square, where two labeled points for two classes are close to diagonally opposite corners. Laplacian regularization would cut the square in half approximately along the other diagonal, Fig.~\ref{fig:noMinCut}, left, while the minimum graph cut would lead to highly asymmetric solution: one predicted label concentrates closely around one of the labeled points, all the rest occupied by the other label. 

However in our experiments we very rarely observe this trend towards asymmetrical solutions. To develop an intuition for why this happens, consider a period in the learning process during which labels channel is silent ($z_t=0$), and $y_t$ is not reaching the limits yet. This is the decisive period, where the label information propagates between the synapse weights, see Fig. \ref{fig:2spirals_zneuron}, bottom left. Then \eqref{eq:ystep} becomes simply $y_t = {\mu \bf w}_t^\top {\bf h}_t$. Assume the input points arrive i.i.d., then so are ${\bf h}_t$ vectors. Then substituting expression for $y_t$ into \eqref{eq:wstep} we can write an expectation for one component of ${\bf w}$:
\begin{equation} \label{eq:exdiff}
    \operatorname{E}( \Delta w_i ) = \eta \Big(\sum_j \mu \operatorname{E}(h_{i} h_{j}) w_j - w_i \Big) = \eta \Big(\mu\sum_j s_{i,j} w_j - w_i \Big),
\end{equation}
where we defined $s_{i,j}\equiv \operatorname{E}(h_{i} h_{j})$, and $\eta=1/(t+1)$. Now $(s_{i,j})$ can be seen as the adjacency matrix of a weighted graph, where vertices are tiling channels, in a manner analogous to the matrix $(S_{i,j})$, appearing in Eq.~\eqref{grLap} in the previous section. The term $\mu\sum_j s_{i,j} w_j$, appearing in the right hand side of Eq.~\eqref{eq:exdiff}, has the effect of ``smoothing" out $\bf w$ in over the tiling channels, in a manner analogous to the effect of the Laplacian penalty presented in \eqref{svmLap}. Essentially, the Laplacian penalty causes the components of $\bf w$  diffuse over the graph \cite{zhu2003semi}. So, in expectation, the evolution of $\bf w$ in our algorithm would share some features with the gradient descent of $\bf w$ to optimize the expression in \eqref{svmLap}. ``Smoothness" of the resulting $\bf w$ over channels translate to smoothness of prediction over input space, thereby reducing the likelihood of extremely imbalanced solutions.

We illustrate this with a simulation experiment on the square in Fig.~\ref{fig:noMinCut}. Ideally, there should be equal number of predicted labels for both classes. We, therefore, measure the imbalance, by looking at fraction associated with the majority class among predictions. This measure of imbalance ranges from 0.5 to 1.0. For each run, we generate 2000 unlabeled sample points uniformly on the square, plus 2 labeled points near the corners. These data were fed to our network with 50 tiling channels and $\mu=10$ in \eqref{eq:zneuronONline2}. For comparison, the output of tiling layer is also used as input to linear classifier with Laplacian regularization. The histogram of results, after 100 such runs, is presented in Fig.~\ref{fig:noMinCut}, right. While indeed the results for our network fluctuate more, compared to those of the Laplacian regularization approach, the extreme imbalances are rare in both approaches.

\section{Conclusion}
We presented a neural network that learns low-dimensional manifolds in the data stream, then learns a classifier in a semi-supervised setting, where only small part of inputs are labeled.  The network operates in an online fashion, producing an output immediately after seeing every input. Weights are updated by a biologically plausible local Hebbian-type rule. We demonstrated the effectiveness of the network in simulations, comparing it with fully supervised online algorithm and with a semi-supervised offline algorithm.

\section*{Acknowledgements}

The authors are grateful to Victor Minden and Mariano Tepper for their insightful comments. We thank Johannes Friedrich, Tiberiu Tesileanu and Charles Windolf for helpful discussions.

%
%






\end{document}